\documentclass{article}
\usepackage{spconf,amsmath,graphicx}
\usepackage{tikz,tikz-cd, pgf}

\usepackage{amssymb}
\usepackage{stmaryrd}
\usepackage{amsfonts}
\usepackage{algorithm}
\usepackage[noend]{algorithmic} %
\usepackage{setspace}
\usepackage{verbatim}
\usepackage{comment} 
\usepackage{url}
\usepackage[colorlinks=true, allcolors=blue]{hyperref}

\newcommand{\etal}{\textit{et al}.}

\title{Reducing Anomaly Detection in Images to Detection in Noise}

\name{Axel Davy$^{\dagger\star}$, Thibaud Ehret$^{\dagger\star}$, Jean-Michel Morel$^\dagger$, Mauricio Delbracio$^{\S}$ \sthanks{Work supported by IDEX  Paris-Saclay IDI 2016, ANR-11-IDEX-0003-02, ONR  grant N00014-17-1-2552,  CNES MISS project, Agencia Nacional de Investigaci\'on e Innovaci\'on (ANII, Uruguay) grant FCE\_1\_2017\_135458, DGA Astrid ANR-17-ASTR-0013-01, DGA ANR-16-DEFA-0004-01, Programme ECOS Sud -- UdelaR - Paris Descartes U17E04, and MENRT.\newline\indent$\star$ Both authors contributed equally to this work}}

\address{$^\dagger$CMLA, ENS Cachan, CNRS, Universit\'e Paris-Saclay, 94235 Cachan, France\\
$^\S$IIE, Universidad de la Rep\'ublica, Uruguay}

\begin{document}

\parbox{\textwidth}{
 © 2018 IEEE.  Personal use of this material is permitted.  Permission from IEEE must be obtained for all other uses, in any current or future media, including reprinting/republishing this material for advertising or promotional purposes, creating new collective works, for resale or redistribution to servers or lists, or reuse of any copyrighted component of this work in other works.\\
 DOI: \href{https://doi.org/10.1109/ICIP.2018.8451059}{https://doi.org/10.1109/ICIP.2018.8451059}
}

\clearpage

\ninept
\maketitle
\begin{abstract}
   Anomaly detectors address the difficult problem of detecting automatically exceptions in an arbitrary background image. Detection methods have been proposed by the thousands because each problem requires a different background model.
By analyzing the existing approaches, we show that the problem can be reduced to detecting anomalies in residual images (extracted from the target image) in which noise and anomalies prevail. Hence, the general and impossible background modeling problem is replaced by simpler noise modeling, and allows the calculation of rigorous thresholds based on the a contrario detection theory. Our approach is therefore unsupervised and works on arbitrary images.   
\end{abstract}

\begin{keywords}
Anomaly detection, Saliency, Self-similarity
\end{keywords}

\section{Introduction}

Anomalies are image regions not conforming  with  the rest of the image. Detecting them is a  challenging image  analysis problem, as there seems to be no straightforward definition of what is (ab)normal for a given image.

Anomalies in images can be high-level or low-level outliers. High-level anomalies are related to the semantic information presented in the scene. For example, human observers immediately detect a person inappropriately dressed  for a given social event. In this work, we focus on the problem of detecting anomalies due to low or mid level rare local patterns present in images. This is an important problem in many industrial, medical or biological applications.

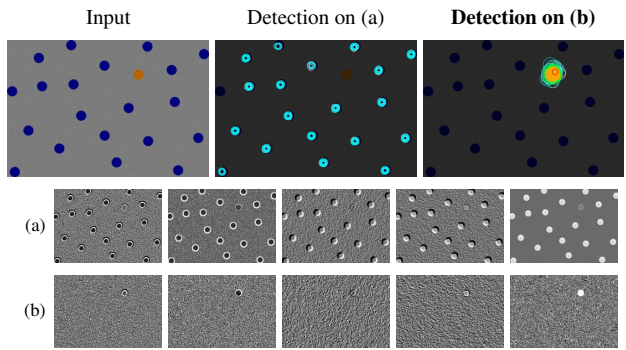
\begin{figure}
\small
  \centering
  \begin{tikzpicture}
  \newlength{\nextfigheighta}
  \newlength{\figwidtha}
  \newlength{\figsepa}
  \setlength{\nextfigheighta}{0cm}
  \setlength{\figwidtha}{0.15\textwidth}
  \setlength{\figsepa}{0.155\textwidth}
\node[anchor=south, inner sep=0] (input) at (0,\nextfigheighta) {\includegraphics[width=\figwidtha]{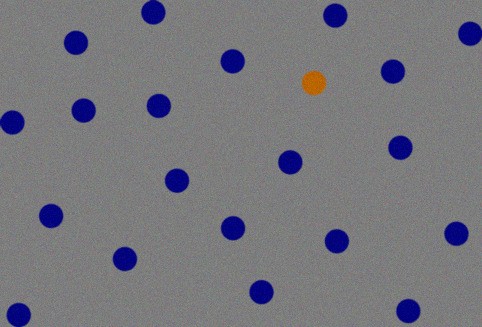}};
\node[anchor=south, inner sep=0] (conv21nod) at (\figsepa,\nextfigheighta){\includegraphics[width=\figwidtha]{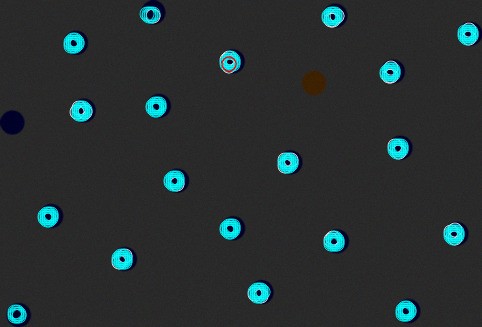}};
\node[anchor=south, inner sep=0] (conv21d)  at (2*\figsepa,\nextfigheighta){\includegraphics[width=\figwidtha]{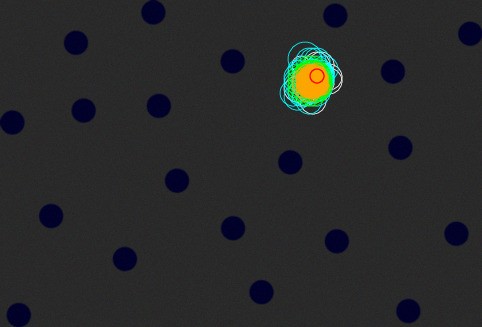}};

\node [anchor=south] at (input.north) {\footnotesize \begin{tabular}{c}\;\\ Input \end{tabular}};
\node [anchor=south] at (conv21nod.north) {\footnotesize {\begin{tabular}{c} Detection on (a) \end{tabular}}};
\node [anchor=south] at (conv21d.north) {\footnotesize \begin{tabular}{c} \;\\ \textbf{Detection on (b)} \end{tabular}};

\setlength{\figwidtha}{0.080\textwidth}
\setlength{\figsepa}{0.085\textwidth}

\addtolength{\nextfigheighta}{-0.75\figsepa}

\node[anchor=south, inner sep=0] (original1) at (0,\nextfigheighta) {\includegraphics[width=\figwidtha]{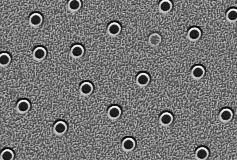}};
\node[anchor=south, inner sep=0] () at (\figsepa,\nextfigheighta){\includegraphics[width=\figwidtha]{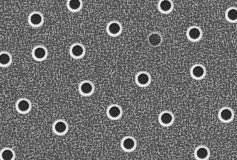}};
\node[anchor=south, inner sep=0] ()  at (2*\figsepa,\nextfigheighta){\includegraphics[width=\figwidtha]{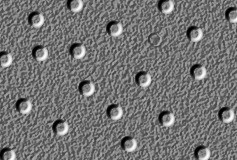}};
\node[anchor=south, inner sep=0] ()  at (3*\figsepa,\nextfigheighta){\includegraphics[width=\figwidtha]{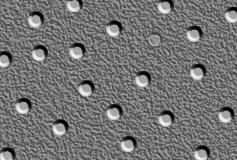}};
\node[anchor=south, inner sep=0] ()  at (4*\figsepa,\nextfigheighta){\includegraphics[width=\figwidtha]{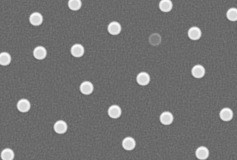}};

\addtolength{\nextfigheighta}{-0.75\figsepa}

\node[anchor=south, inner sep=0] (residual1) at (0,\nextfigheighta) {\includegraphics[width=\figwidtha]{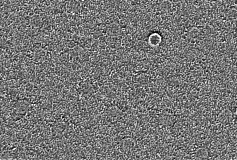}};
\node[anchor=south, inner sep=0] () at (\figsepa,\nextfigheighta){\includegraphics[width=\figwidtha]{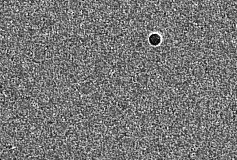}};
\node[anchor=south, inner sep=0] ()  at (2*\figsepa,\nextfigheighta){\includegraphics[width=\figwidtha]{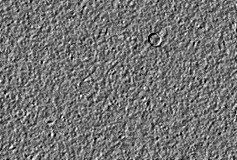}};
\node[anchor=south, inner sep=0] ()  at (3*\figsepa,\nextfigheighta){\includegraphics[width=\figwidtha]{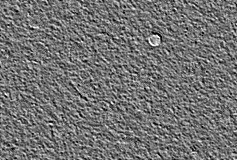}};
\node[anchor=south, inner sep=0] ()  at (4*\figsepa,\nextfigheighta){\includegraphics[width=\figwidtha]{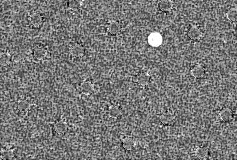}};

\node [anchor=east] at (original1.west) {\scriptsize (a)};
\node [anchor=east] at (residual1.west) {\scriptsize (b)};

  \end{tikzpicture}
\caption{Image anomalies are successfully detected by removing all self-similar content and then looking for structure in the residual noise. 
Top row: left, an image with a color anomaly (the red dot); middle, detections obtained from top five principal components of CNN features shown in (a); right, detections on features shown in (b), obtained after removing the self-similar content. Cyan corresponds to good detection and orange extremely salient detection.
}
 
\label{fig:conv_denoising}
\end{figure}

We introduce in  this  paper an unsupervised method for detecting anomalies in an arbitrary image. The method does not rely on a training dataset of normal or abnormal images, neither on any other prior knowledge about the image statistics.  It directly detects anomalies with respect to residual images estimated solely from the image itself.  We only use a generic, qualitative background image model: we assume that anything that repeats in an image is \textit{not} an anomaly.
In a nutshell, our method removes from the image its  self-similar content (considered as being normal).  The residual is modeled as colored Gaussian noise, but still contains the anomalies according to their definition: they do not repeat.  

Detecting anomalies in noise is far easier and can be made rigorous and unsupervised by the \emph{a-contrario} theory~\cite{desolneux2007gestalt} which is a probabilistic formalization of the \emph{non-accidentalness} principle~\cite{lowe1985perceptual}.  The \emph{a-contrario} framework has produced impressive results in many different detection or estimation computer vision tasks, such as, segment detection~\cite{grompone2010lsd},  spots detection \cite{grosjean2009contrario}, vanishing points detection~\cite{lezama2014finding}, mirror-symmetry detection~\cite{patraucean2013detection}, among others. The fundamental property of the \emph{a-contrario} theory is that it provides a way for automatically computing detection thresholds that yield a control on the number of false alarms (NFA). It favorably replaces the  usual  p-value when multiple testing  is  involved. 
It follows that not  only  one can detect anomalies in arbitrary images without complex modeling, but in addition the anomalies are associated an NFA which is often very small and therefore offers a strong guarantee of the  validity of the detection.
We shall show detections performed directly on the image residual, or alternatively on residuals extracted from dense low and mid-level features of the VGG neural net~\cite{simonyan2014very}.

The paper is organized as follows. Section~\ref{sec:relatedWork} discusses previous work while
Section~\ref{sec:method} explains the proposed method and its implementation.  Section~\ref{sec:experiments} presents results of the proposed method on real/synthetic data, and a comparison to other state-of-the-art anomaly detectors. We finally close in Section~\ref{sec:conclusions}.

\vspace{-.5em}
\section{Related Work}
\label{sec:relatedWork} 
\vspace{-.5em}

The 2009 review \cite{chandola2009anomaly} examining about 400 papers on anomaly detection  considered allegedly all existing techniques and application fields. It is fairly well completed by the more recent \cite{pimentel2014review} review. 
These reviews agree that classification techniques like SVM can be discarded, because anomalies are generally not observed in sufficient number and lack statistical coherence. 
There are exceptions like the recent method~\cite{ding2014experimental} which defines anomalies as rare events that cannot be learned, but after estimating a background density model, the right detection thresholds are nevertheless learned from anomalies.  
A broad related literature exists on saliency measures, for which learning from average fixation maps by humans is possible \cite{tavakoli2011fast}. 
Saliency detectors try to mimic the human visual perception and in general introduce semantic prior knowledge (e.g., face detectors). This approach works particularly well with neural networks trained on a base of detect/non-detect  with ground truth obtained by for example, gaze trackers\cite{huang2015salicon}. %

Anomaly detection has been generally handled as a ``one class'' classification problem.   In~\cite{markou2003noveltyA} authors concluded that most research on anomaly detection was driven by modeling background data distributions, to estimate the probability that test data do not belong to such distributions \cite{grosjean2009contrario,honda2001finding,goldman2004anomaly,aiger2010phase}. %
Autoencoders neural networks can be used to model background~\cite{An2016,Schlegl2017}. The general idea is to compute the norm between the input and a reconstruction of the input. Another successful background based method is the detection of anomalies in periodic patterns of textile \cite{tsai2003automated,perng2010novel}. %
In~\cite{itti1998model,murray2011saliency}, center surround detectors based on color, orientation and intensity filters are combined to produce a final saliency map. 
Detection in image and video is also done in \cite{gao2008discriminant} with center-surround saliency detectors which stem from \cite{itti2000saliency} adopting similar image features. In~\cite{honda2001finding}, the main idea is to estimate the probability of a region conditioned on the surroundings.  
A more recent non parametric trend is to learn a sparse dictionary representing the background (i.e., \emph{normality}) and to characterize outliers by their non-sparsity \cite{margolin2013makes,boracchi2014novelty,elhamifar2012see,adler2015sparse,carrera2015detecting}. %

The self-similarity principle has been successfully used in many different applications~\cite{efros1999texture,buades2005non}. %
The basic assumption of this generic background model, is that in normal data, features are densely clustered. Anomalies instead occur far from their closest neighbors.  
This idea is implemented by clustering (anomalies being detected as far away from the centroid of their own cluster), or by simple rarity measurements based on nearest neighbor search (NN)~\cite{boiman2007detecting,seo2009static,goferman2012context}. %

Background probabilistic modeling is  powerful when images belong to a restricted class of homogeneous objects, like textiles.
But, regrettably, this method is nearly impossible to apply on generic images. 
Similarly, background reconstruction models based on CNNs are restrictive and do not rely on provable detection thresholds. 
Center-surround contrast methods are successful for saliency enhancement, but lack a formal detection mechanism. 
Being universal, the sparsity and the self-similarity models are tempting and thriving. But again, they lack a rigorous detection mechanism, because they work on a feature  space that is not easily modeled.

We propose to benefit of the above methods while avoiding their mentioned limitations. To this aim, we do construct a probabilistic background model, but it is applied to a new feature image that we call the \emph{residual}.  
This residual is obtained by computing the difference between a self-similar version of the target image and the target itself.
Being not self-similar, this background is akin to a colored noise.  Hence a hypothesis test can be applied, and more precisely multiple hypothesis testing (also called \textit{a contrario} method), as proposed in~\cite{grosjean2009contrario}. 
In that way, we present a general and simple method that is universal and detects anomalies by a rigorous threshold. It does not require learning, and it is easily made multiscale.

\vspace{-.5em}
\section{Method}
\label{sec:method}
\vspace{-.5em}

Our method is built  on two main blocks: a removal of the self-similar image component, and a simple statistical detection test on the residual based on the \textit{a contrario} framework.

\vspace{-.5em}
\subsection{Construction  of  the  residual image}
\label{denoising}
\vspace{-.5em}

The proposed self-similarity based background subtraction is inspired from patch-based non-local denoising algorithms, where the estimate is done from a set of similar patches~\cite{buades2005non}. This search is generally performed locally around each patch \cite{dabov2007image,buades2005non} to keep computational cost low and to avoid noise overfitting.  
The main difference with non-local denoisers is that  we \textit{forbid} local comparisons. The nearest neighbor search is performed \textit{outside a  square region surrounding each query patch}. This square region is defined as the union of all the patches intersecting the query patch. Otherwise any anomaly with some internal structure might be considered a valid structure.  What matters is that  the event represented by the anomaly is unique, and this is checked away from it.  

For each patch $P$ in the image the $n$ most similar patches denoted by $P_i$ are searched and averaged to give a self-similar estimate,
\begin{equation}
\hat{P} = \frac{1}{Z}\sum_{i=1}^{n} \exp \left(-\frac{ \|P - P_i\|_2^2 }{h^2}\right) P_i
\label{eq:nlmeans}
\end{equation}
where $Z=\sum_{i=1}^{n} \exp \left(-\frac{ \|P - P_i\|_2^2 }{h^2}\right)$ is a normalizing constant, and $h$ is a  parameter. 

Since each pixel belongs to several different patches, they will therefore receive several distinct estimates that can be averaged. Algorithm \ref{algo:model} gives a generic pseudocode for this process, which ends with the generation of a residual image $r(u)$ allegedly containing only noise and the anomalies (see Figure~\ref{fig:conv_denoising}). The intuition is that it is much easier to detect anomalies in $r(u)$ than in $u$.

\begin{algorithm}[t]
\caption{Computation of the unstructured residual}

\begin{spacing}{1.0}
\begin{algorithmic}[1]
\REQUIRE Multichannel Image $u$, $n$ the number of nearest neighbors
\ENSURE Model $\hat{u}$ of $u$ based on $\mathcal{D}$, residual $r(u)=\hat{u}-u$.
\FORALL{Multichannel patch $P$ of $u$}
\STATE Compute $n$ near.neigh. $\{P_i\}$ of $P$ \textbf{(outside square region)}.
\STATE Reconstruct the patch (using \eqref{eq:nlmeans})
\ENDFOR
\FORALL{pixels $j$ in $u$}
\STATE $\hat{u}(j) = \frac{\sum_{i \in \{s|j \in W_s, s \in \llbracket 1, N \rrbracket\}}^{} \hat{P}_i(j)}{\#\{s|j \in W_s, s \in \llbracket 1, N \rrbracket\}}$
\ENDFOR
\end{algorithmic}
\end{spacing}

\textbf{Notation convention.} $W_s:$ set of pixels in the patch centered at $s$. $\hat{P}_i(j):$ value at pixel $j$ of the reconstructed patch centered at $i$.
\label{algo:model}
\end{algorithm}

\subsection{Statistical detection by  the \textbf{\textit{a contrario}} approach}

Our goal is to detect structure in the residual image $r(u)=\hat{u}-u$.   We are in a much better situation modeling $r(u)$ than $u$.  Indeed, contrarily to $u$, $r(u)$ is by construction \emph{unstructured} and akin to a colored noise (as illustrated in Fig.~\ref{fig:conv_denoising}). In what follows we  assume that $r(u)$ is a spatial stationary random process and follow~\cite{grosjean2009contrario}, who proposed automatic detection thresholds in any colored Gaussian noise.
 
Given a set of random variables $(X_i)_{i\in[|1,N|]}$
a function $f$ is called an NFA if it guarantees a bound on the expectation of its number of false alarms  under the null-hypothesis, namely,
$\forall{\epsilon>0}, \mathbb{E}[\#\{i, f(i, X_i) \le \epsilon\}] \le \epsilon$.
In other words, thresholding all the $f(i, X_i)$ by $\epsilon$ should give up to $\epsilon$ false alarms when $(X_i)_{i\in[|1,N|]}$ verifies the null-hypothesis.
In our case, we consider 
\begin{equation}
f(i, \mathbf{x}) = N \mathbb{P}(|X_i| \ge |x_i|), 
\label{eq:nfa}
\end{equation}
Where $i$ index among the $N$ executed tests (detailed below), $X_i$ is a random variable distributed as the residual at position $i$, and $x_i$ the actual measured value (pixel or feature value) at position $i$. The null-hypothesis is that the residual, represented by $(X_i)_{i\in[|1,N|]}$, verifies that each $X_i$ follows a standard normal distribution. Independence is not required.

\noindent \textbf{Residual distribution.}
In practice the distribution of the residual $r(u)$ is not necessarily Gaussian. A careful study of the residual distribution lead us to consider that it follows a generalized Gaussian distribution (GCD). We approximately estimate the GCD parameters, and then apply a non-linear mapping to make it normally distributed.

\vspace{.25em}  \noindent \textbf{Choice of NFA.}
The choice of the NFA given in~\eqref{eq:nfa} enables to detect anomalies in both tails of the Gaussian distribution (i.e., very bright or very dark spots).
To detect anomalies of all sizes, the detection is carried out independently at $N_\text{scales}$ scales computed from the residual at the original resolution (by Gaussian subsampling of factor two). Let us denote by $\Omega_s$ the set of pixels in the residual image at scale $s$ having $N_\text{feat}$ number of features.
When working with colored noise, Grosjean and Moisan~\cite{grosjean2009contrario} propose to convolve the noise with a measure kernel to detect spots of a certain size. This corresponds to the generation of new image features $\bar{r}(u) = r(u) \ast K$,
where $K$ is a disk of a given radius. This idea is used in our framework, where the residual is convolved with kernels of small sizes. Since we apply the detection at all dyadic scales, the tested radii are limited to a small set of $N_\text{kernel}$  values (1,2 to 3) at each scale. Because the residual is assumed to be a stationary Gaussian field, the result after filtering is also Gaussian. The variance is estimated and the filtered residuals are normalized to have unit variance. This is the input to the NFA~\eqref{eq:nfa} computation (i.e., $\textbf{x}_i$). Thus, the inputs to the detection phase are multi-channel images of different scales, where each pixel channel, representing a given feature, follows a standard normal distribution.

Then,  the number of tests is
$
N = N_\text{kernel} \cdot N_\text{feat} \cdot \sum_{i=0}^{N_\text{scales}-1} |\Omega_s|.
 $

\vspace{-.5em}

\subsection{Choice of the image features}
\vspace{-.5em}
\label{features}

Anomaly detectors work either directly on image pixels or on some feature space but the detection in the residual, which is akin to unstructured noise, is fairly independent of the choice of the features. We used with equal success the raw image color pixels, or some intermediate feature representation extracted from the  VGG convolutional neural network~\cite{simonyan2014very}. %
To compress the dynamical range of the feature space we apply a square root function to the network features.

In order to reduce the feature space dimension, we compute the principal components (PCA) and keep only the first five. This is done per input image independently.

\noindent \textbf{Parameters.}
The main method parameter is the number of allowed false alarms in the statistical test. In all presented experiments, we set NFA=$10^{-2}$.  Hence, an anomaly is detected at pixel $\mathbf{x}$ in channel $i$ iff the NFA function $f(i, \mathbf{x})$ is below $\epsilon=10^{-2}$. This implies a (theoretical) expectation of less than $10^{-2}$ ``casual'' detection per image under the null hypothesis that  the residual image is noise. 
 Obviously the lower the NFA the better. Most anomalies  have a much lower NFA.
For the basic method working on image pixels we used two disks of radius one and two, while for the neural network features, we add a third disk of radius three.
The number of scales is set to $N_{scale}=4$ in all tests. The patch size in Alg.~\ref{algo:model} is $8\! \times\! 8\! \times\! 3$ for the pixels variant, while when using neural nets features, we use a patch size of $5\! \times\! 5\! \times\! 5$. The number of nearest patches is always set to $n=16$, and $h=10$.  Results presented herein use the outputs from VGG-19 layers \verb!conv1_1!, \verb!conv2_1! and \verb!conv3_1!. %

\vspace{-.5em}

\section{Experiments}
\label{sec:experiments}
\vspace{-.5em}
\begin{figure*}[th]
  \centering
  \begin{tikzpicture}
  \newlength{\nextfigheight}
  \newlength{\figwidth}
  \newlength{\figsep}
  \setlength{\nextfigheight}{0cm}
  \setlength{\figwidth}{0.105\textwidth}
  \setlength{\figsep}{0.11\textwidth}
    \node[anchor=south, inner sep=0] (example1) at (0,\nextfigheight) {\includegraphics[width=\figwidth]{Experiments/Toy/color}};
\node[anchor=south, inner sep=0] (nonn1) at (\figsep,\nextfigheight){\includegraphics[width=\figwidth]{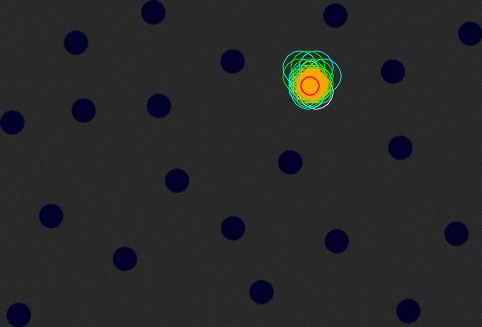}};
\node[anchor=south, inner sep=0] (conv111)  at (2*\figsep,\nextfigheight){\includegraphics[width=\figwidth]{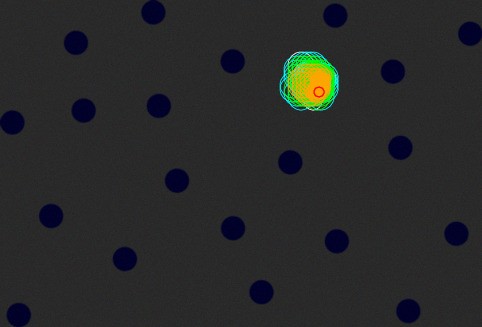}};
\node[anchor=south, inner sep=0] (conv211)  at (3*\figsep,\nextfigheight){\includegraphics[width=\figwidth]{Experiments/Toy/detections_conv21_color}};
\node[anchor=south, inner sep=0] (conv311)  at (4*\figsep,\nextfigheight){\includegraphics[width=\figwidth]{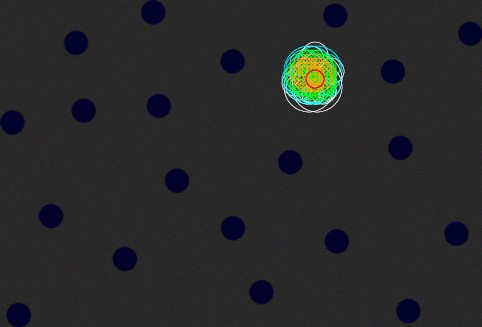}};
\node[anchor=south, inner sep=0] (salicon1)  at (8*\figsep,\nextfigheight){\includegraphics[width=\figwidth]{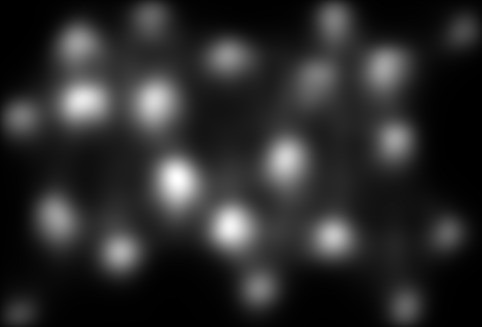}};
\node[anchor=south, inner sep=0] (itti1) at (6*\figsep,\nextfigheight){\includegraphics[width=\figwidth]{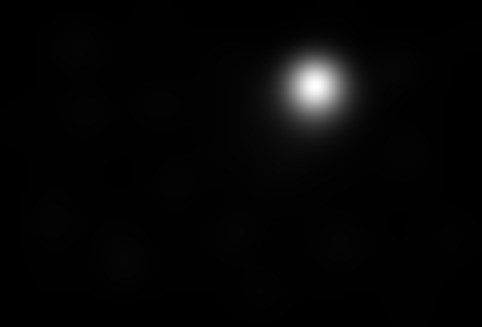}};
\node[anchor=south, inner sep=0] (cohen1) at (5*\figsep,\nextfigheight){\includegraphics[width=\figwidth]{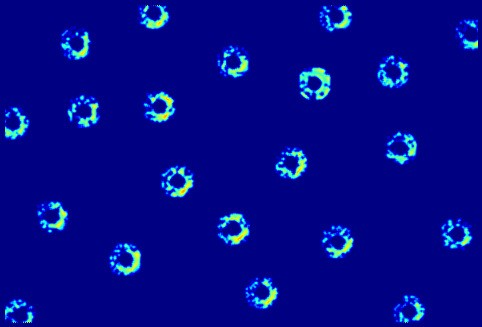}};
\node[anchor=south, inner sep=0] (drfi1) at (7*\figsep,\nextfigheight){\includegraphics[width=\figwidth]{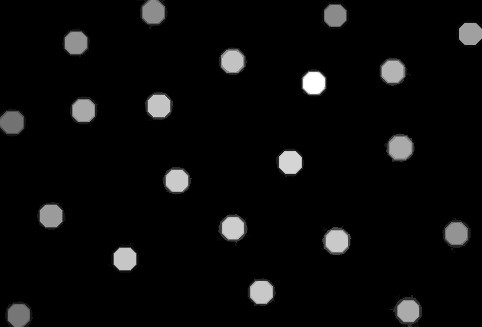}};
    \node [anchor=south] at (example1.north) {\scriptsize Input};
    \node [anchor=south] at (nonn1.north) {\scriptsize \vphantom{p}\texttt{pixels}\vphantom{p}};
    \node [anchor=south] at (conv111.north) {\scriptsize \vphantom{p}\texttt{conv1\_1}\vphantom{p}};
    \node [anchor=south] at (conv211.north) {\scriptsize \vphantom{p}\texttt{conv2\_1}\vphantom{p}};
    \node [anchor=south] at (conv311.north) {\scriptsize \vphantom{p}\texttt{conv3\_1}\vphantom{p}};
    \node [anchor=south] at (salicon1.north) {\scriptsize \vphantom{p}SALICON \cite{huang2015salicon}\vphantom{p}};
    \node [anchor=south] at (itti1.north) {\scriptsize \vphantom{p}Itti \textit{et al.}\cite{itti1998model}\vphantom{p}};
    \node [anchor=south] at (cohen1.north) {\scriptsize \vphantom{p}Mishne \!-\! Cohen \cite{mishne2013multiscale}\vphantom{p}};
    \node [anchor=south] at (drfi1.north) {\scriptsize \vphantom{p}DRFI \cite{jiang2013salient}\vphantom{p}};

\addtolength{\nextfigheight}{-0.68\figsep}

\node[anchor=south, inner sep=0] (example2) at (0,\nextfigheight) {\includegraphics[width=\figwidth]{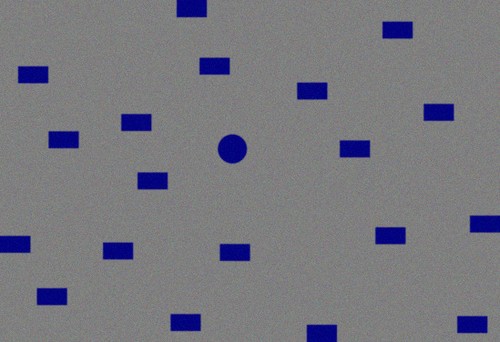}};
\node[anchor=south, inner sep=0] (nonn2) at (\figsep,\nextfigheight){\includegraphics[width=\figwidth]{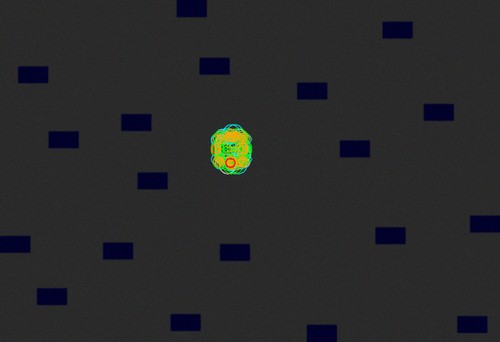}};
\node[anchor=south, inner sep=0] (conv112)  at (2*\figsep,\nextfigheight){\includegraphics[width=\figwidth]{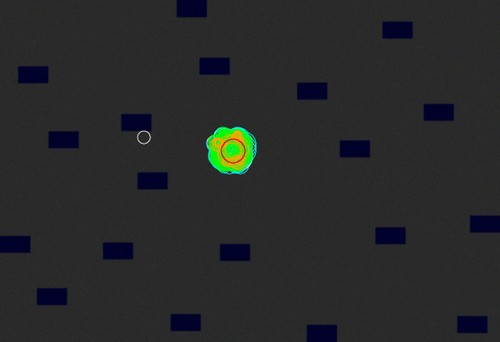}};
\node[anchor=south, inner sep=0] (conv212)  at (3*\figsep,\nextfigheight){\includegraphics[width=\figwidth]{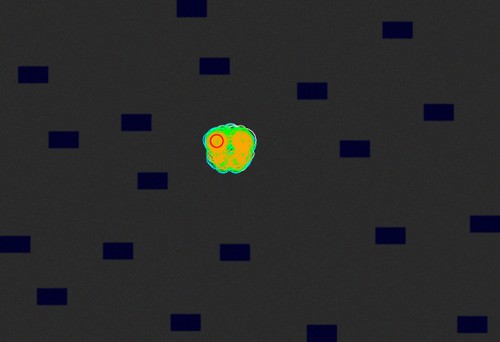}};
\node[anchor=south, inner sep=0] (conv312)  at (4*\figsep,\nextfigheight){\includegraphics[width=\figwidth]{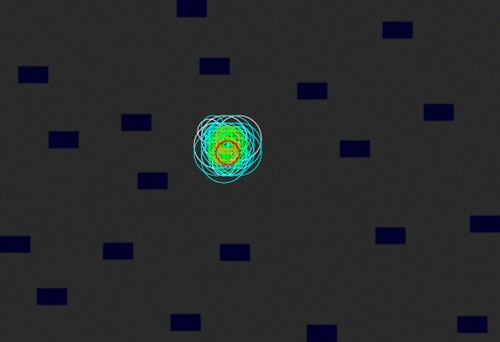}};
\node[anchor=south, inner sep=0] (salicon2)  at (8*\figsep,\nextfigheight){\includegraphics[width=\figwidth]{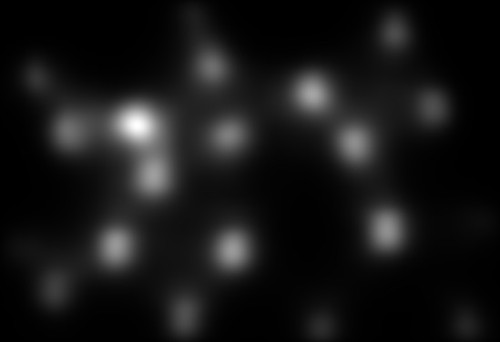}};
\node[anchor=south, inner sep=0] (itti2) at (6*\figsep,\nextfigheight){\includegraphics[width=\figwidth]{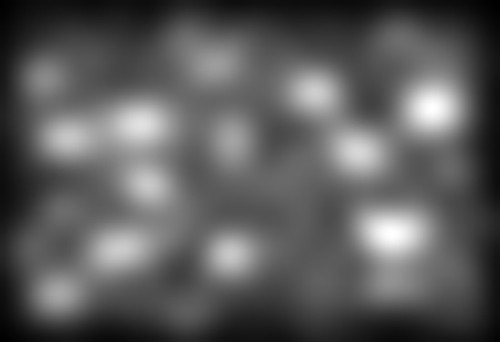}};
\node[anchor=south, inner sep=0] (cohen2) at (5*\figsep,\nextfigheight){\includegraphics[width=\figwidth]{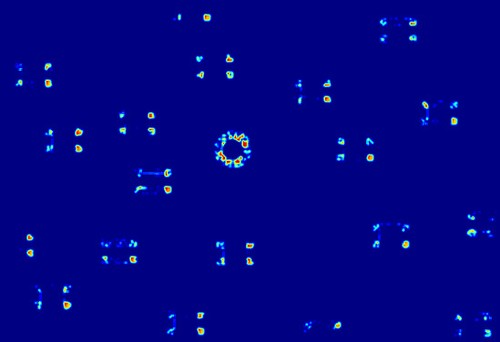}};
\node[anchor=south, inner sep=0] (drfi2) at (7*\figsep,\nextfigheight){\includegraphics[width=\figwidth]{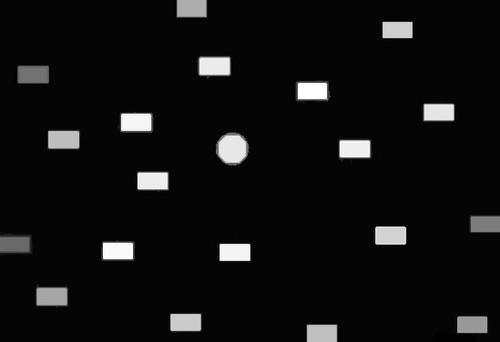}};

\addtolength{\nextfigheight}{-0.61\figsep}

\node[anchor=south, inner sep=0] (example5) at (0,\nextfigheight) {\includegraphics[width=\figwidth]{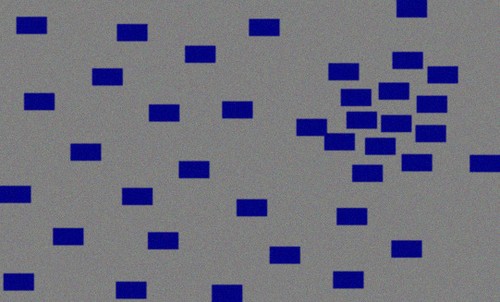}};
\node[anchor=south, inner sep=0] (nonn5) at (\figsep,\nextfigheight){\includegraphics[width=\figwidth]{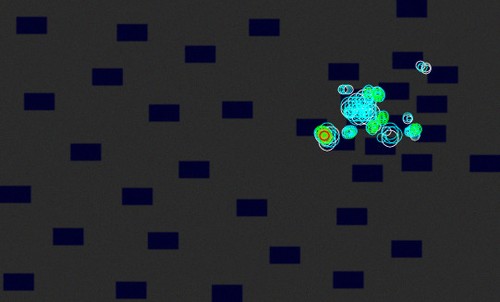}};
\node[anchor=south, inner sep=0] (conv115)  at (2*\figsep,\nextfigheight){\includegraphics[width=\figwidth]{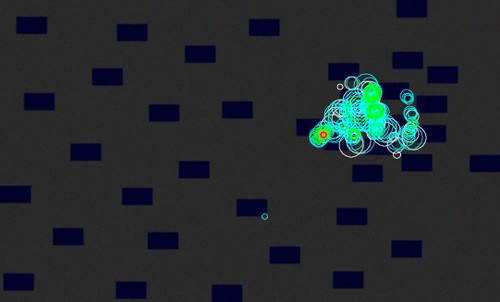}};
\node[anchor=south, inner sep=0] (conv215)  at (3*\figsep,\nextfigheight){\includegraphics[width=\figwidth]{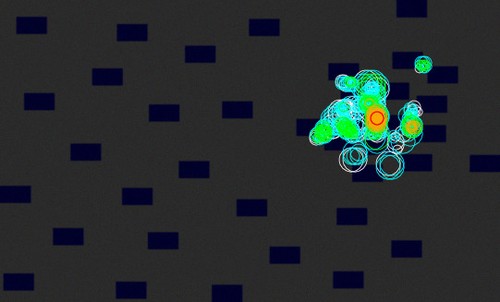}};
\node[anchor=south, inner sep=0] (conv315)  at (4*\figsep,\nextfigheight){\includegraphics[width=\figwidth]{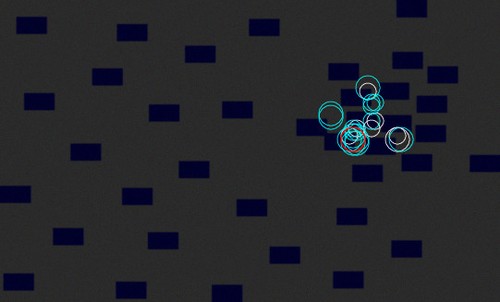}};
\node[anchor=south, inner sep=0] (salicon5)  at (8*\figsep,\nextfigheight){\includegraphics[width=\figwidth]{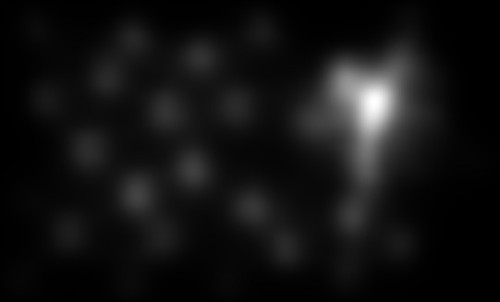}};
\node[anchor=south, inner sep=0] (itti5) at (6*\figsep,\nextfigheight){\includegraphics[width=\figwidth]{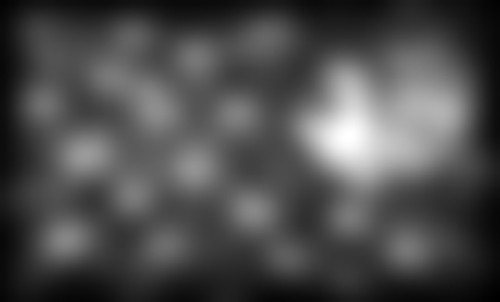}};
\node[anchor=south, inner sep=0] (cohen5) at (5*\figsep,\nextfigheight){\includegraphics[width=\figwidth]{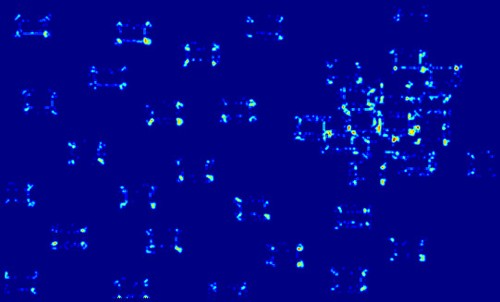}};
\node[anchor=south, inner sep=0] (drfi5) at (7*\figsep,\nextfigheight){\includegraphics[width=\figwidth]{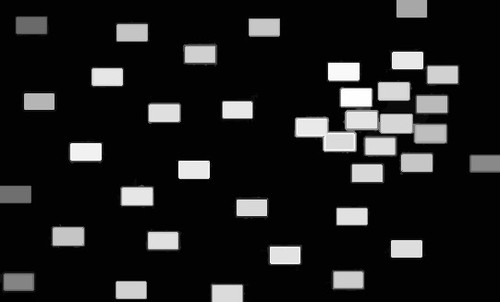}};

\addtolength{\nextfigheight}{-0.72\figsep}
\node[anchor=south, inner sep=0] (example6) at (0,\nextfigheight) {\includegraphics[clip,trim=0 70 0 70,width=\figwidth]{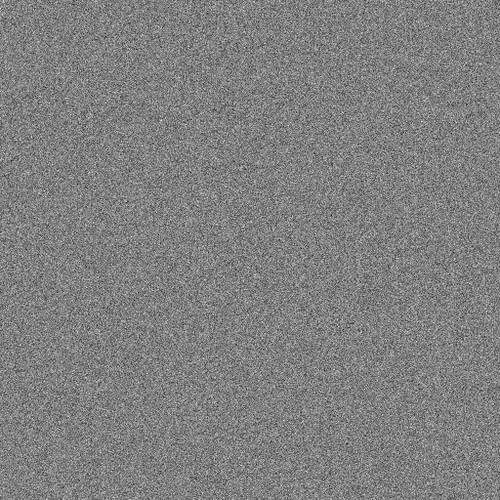}};
\node[anchor=south, inner sep=0] (nonn6) at (\figsep,\nextfigheight){\includegraphics[clip,trim=0 70 0 70,width=\figwidth]{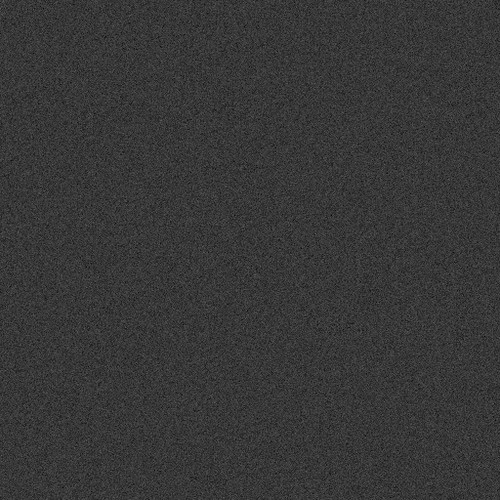}};
\node[anchor=south, inner sep=0] (conv116)  at (2*\figsep,\nextfigheight){\includegraphics[clip,trim=0 70 0 70, width=\figwidth]{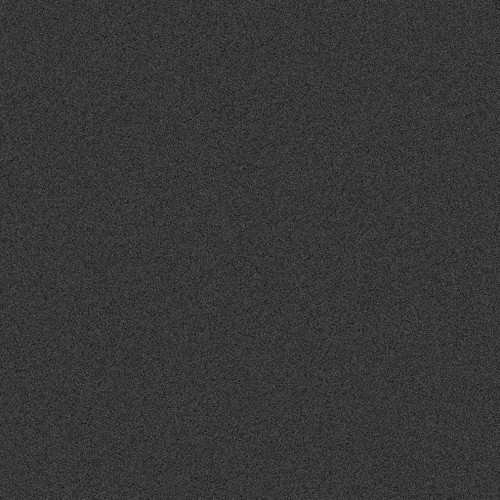}};
\node[anchor=south, inner sep=0] (conv216)  at (3*\figsep,\nextfigheight){\includegraphics[clip,trim=0 70 0 70,width=\figwidth]{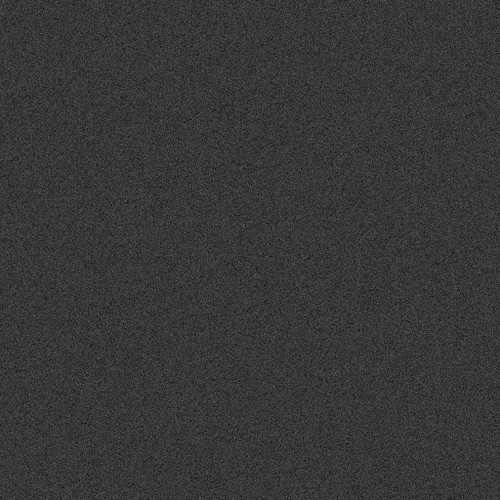}};
\node[anchor=south, inner sep=0] (conv316)  at (4*\figsep,\nextfigheight){\includegraphics[clip,trim=0 70 0 70,width=\figwidth]{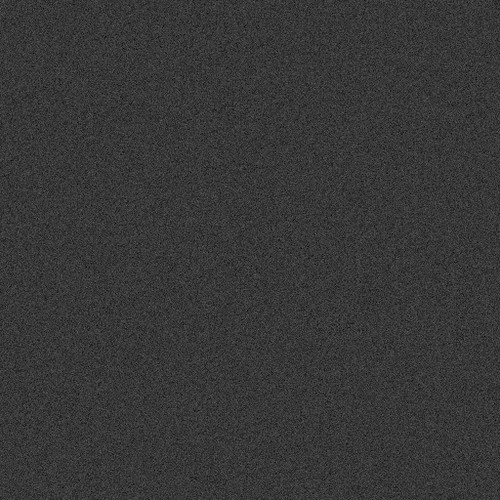}};
\node[anchor=south, inner sep=0] (salicon6)  at (8*\figsep,\nextfigheight){\includegraphics[clip,trim=0 70 0 70,width=\figwidth]{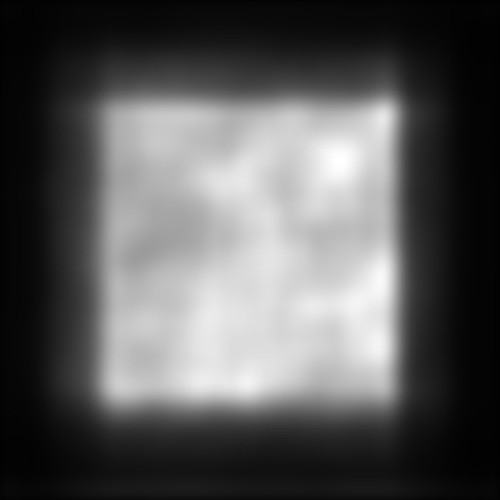}};
\node[anchor=south, inner sep=0] (itti6) at (6*\figsep,\nextfigheight){\includegraphics[clip,trim=0 70 0 70,width=\figwidth]{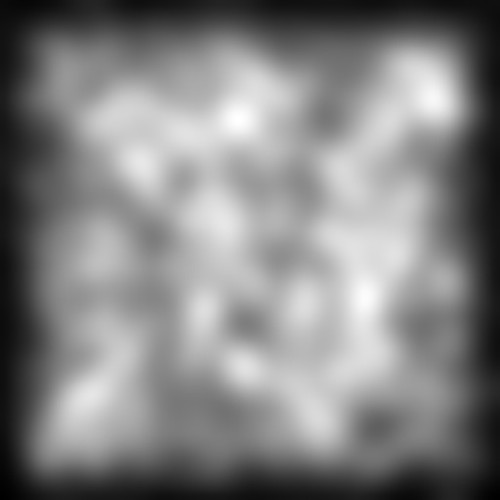}};
\node[anchor=south, inner sep=0] (cohen6) at (5*\figsep,\nextfigheight){\includegraphics[clip,trim=0 70 0 70,width=\figwidth]{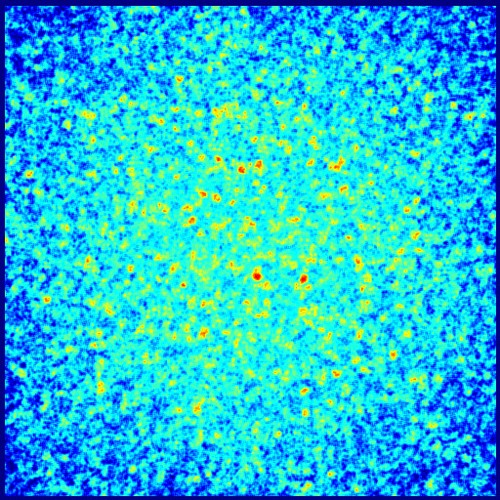}};
\node[anchor=south, inner sep=0] (drfi6) at (7*\figsep,\nextfigheight){\includegraphics[clip,trim=0 70 0 70,width=\figwidth]{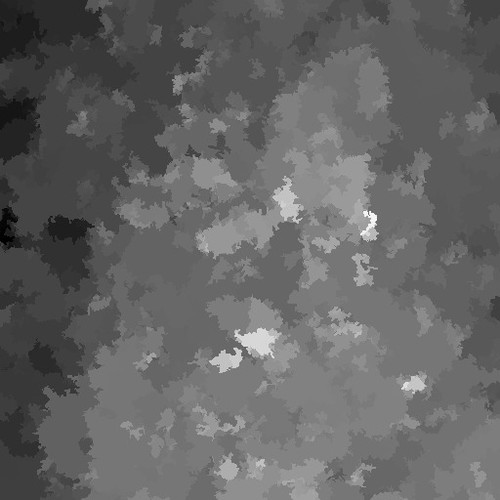}};

 \end{tikzpicture}
 
 \vspace{.15em}
  
\begin{tikzpicture}
  \newlength{\nextfigheightd}
  \newlength{\figwidthd}
  \newlength{\figsepd}
  \setlength{\nextfigheightd}{0cm}
  \setlength{\figwidthd}{0.105\textwidth}
  \setlength{\figsepd}{0.11\textwidth}

    \node[anchor=south, inner sep=0] (example1) at (0,\nextfigheightd) {\includegraphics[width=\figwidthd]{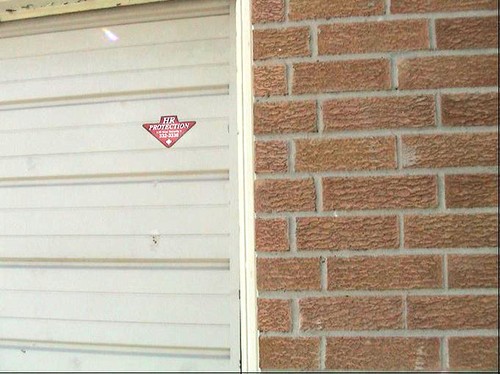}};
\node[anchor=south, inner sep=0] (nonn1) at (\figsepd,\nextfigheightd){\includegraphics[width=\figwidthd]{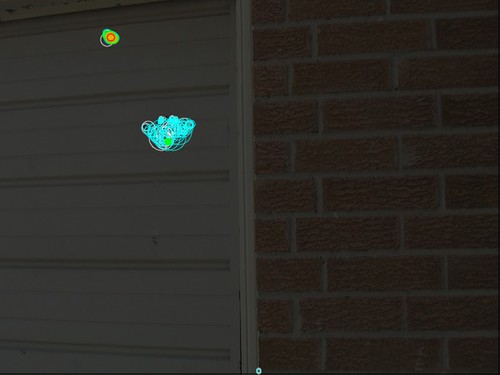}};
\node[anchor=south, inner sep=0] (conv111)  at (2*\figsepd,\nextfigheightd){\includegraphics[width=\figwidthd]{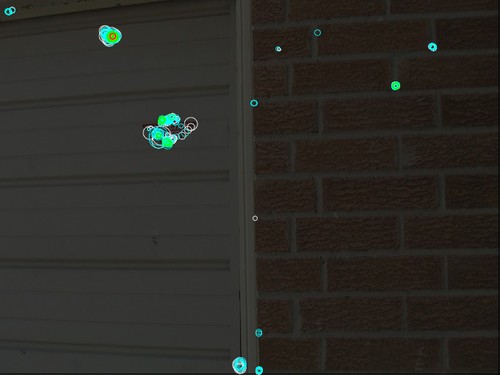}};
\node[anchor=south, inner sep=0] (conv211)  at (3*\figsepd,\nextfigheightd){\includegraphics[width=\figwidthd]{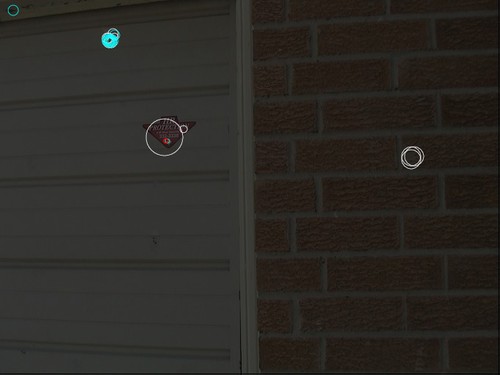}};
\node[anchor=south, inner sep=0] (conv311)  at (4*\figsepd,\nextfigheightd){\includegraphics[width=\figwidthd]{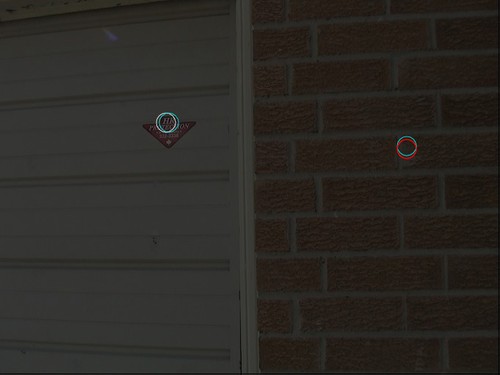}};
\node[anchor=south, inner sep=0] (salicon1)  at (8*\figsepd,\nextfigheightd){\includegraphics[width=\figwidthd]{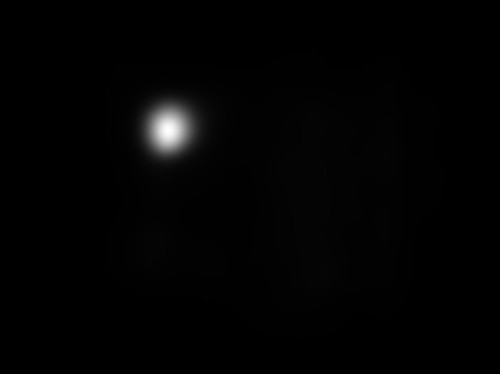}};
\node[anchor=south, inner sep=0] (itti1) at (6*\figsepd,\nextfigheightd){\includegraphics[width=\figwidthd]{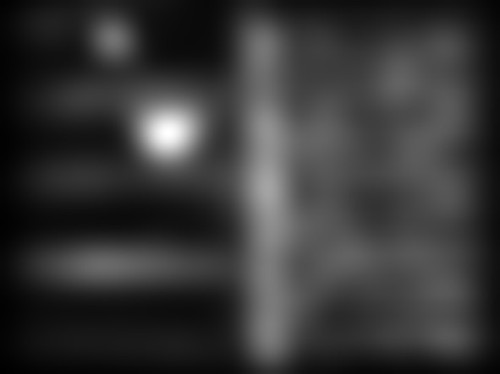}};
\node[anchor=south, inner sep=0] (cohen1) at (5*\figsepd,\nextfigheightd){\includegraphics[width=\figwidthd]{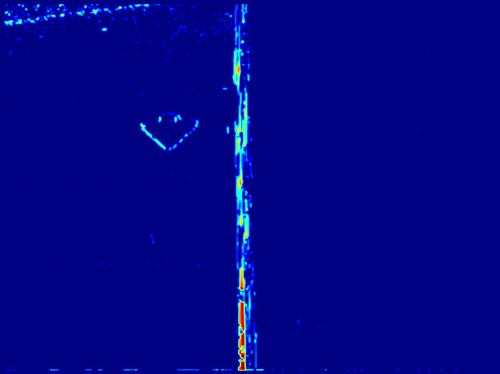}};
\node[anchor=south, inner sep=0] (drfi1) at (7*\figsepd,\nextfigheightd){\includegraphics[width=\figwidthd]{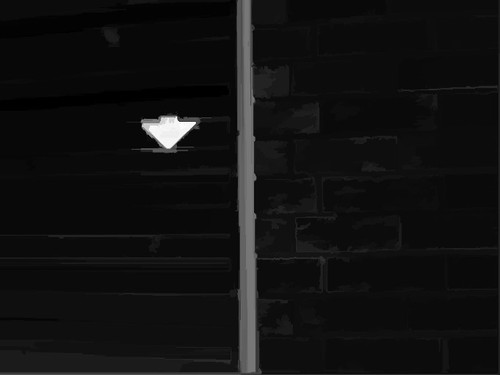}};

\addtolength{\nextfigheightd}{-0.75\figsepd}

\node[anchor=south, inner sep=0] (example2) at (0,\nextfigheightd) {\includegraphics[width=\figwidthd]{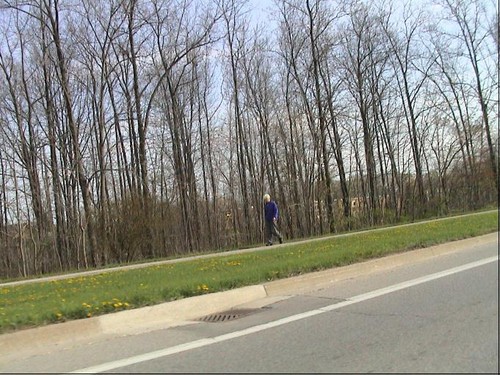}};
\node[anchor=south, inner sep=0] (nonn2) at (\figsepd,\nextfigheightd){\includegraphics[width=\figwidthd]{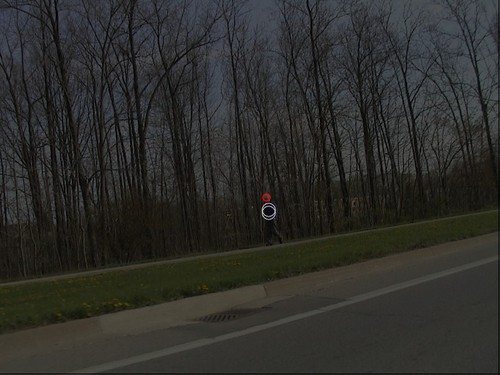}};
\node[anchor=south, inner sep=0] (conv112)  at (2*\figsepd,\nextfigheightd){\includegraphics[width=\figwidthd]{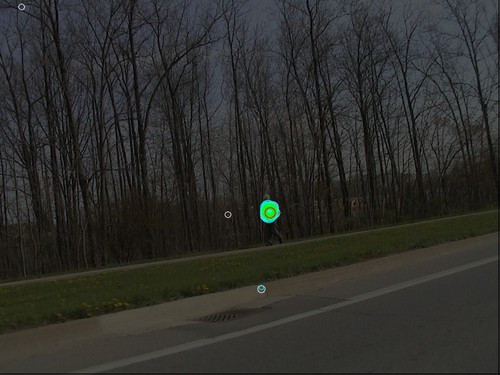}};
\node[anchor=south, inner sep=0] (conv212)  at (3*\figsepd,\nextfigheightd){\includegraphics[width=\figwidthd]{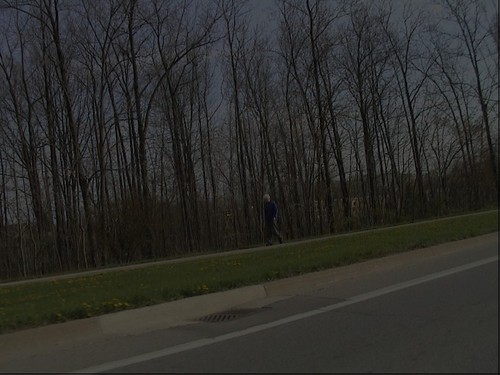}};
\node[anchor=south, inner sep=0] (conv312)  at (4*\figsepd,\nextfigheightd){\includegraphics[width=\figwidthd]{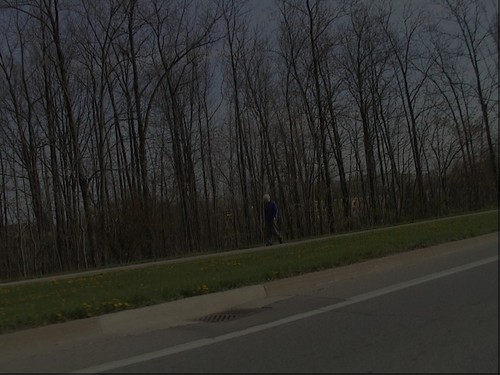}};
\node[anchor=south, inner sep=0] (salicon2)  at (8*\figsepd,\nextfigheightd){\includegraphics[width=\figwidthd]{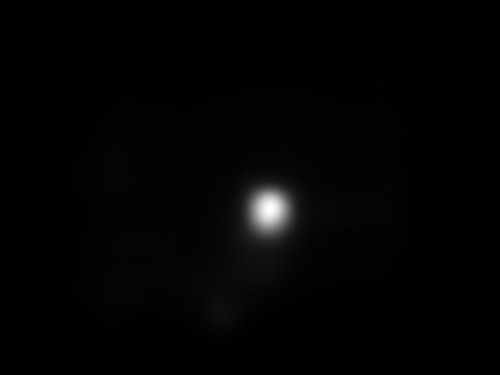}};
\node[anchor=south, inner sep=0] (itti2) at (6*\figsepd,\nextfigheightd){\includegraphics[width=\figwidthd]{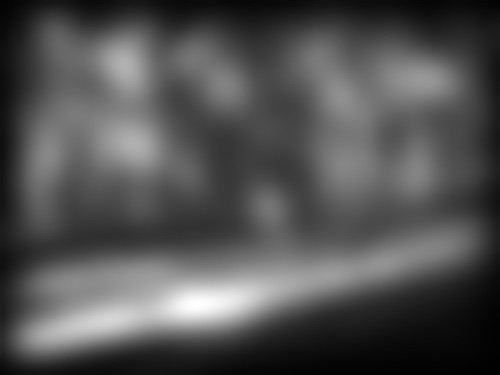}};
\node[anchor=south, inner sep=0] (cohen2) at (5*\figsepd,\nextfigheightd){\includegraphics[width=\figwidthd]{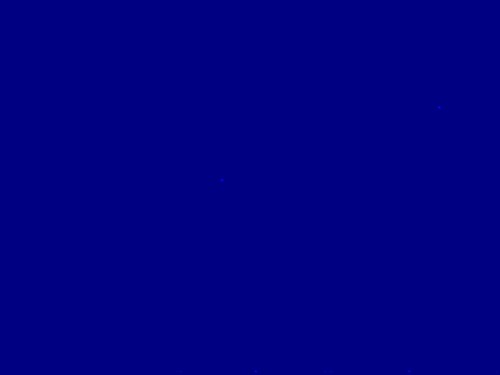}};
\node[anchor=south, inner sep=0] (drfi2) at (7*\figsepd,\nextfigheightd){\includegraphics[width=\figwidthd]{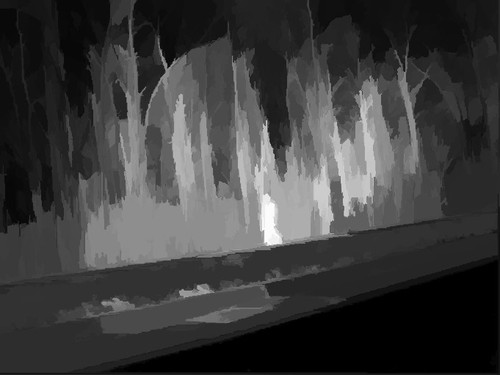}};

\addtolength{\nextfigheightd}{-0.9\figsepd}

\node[anchor=south, inner sep=0] (example3) at (0,\nextfigheightd) {\includegraphics[clip,trim=0 15 0 15,width=\figwidthd]{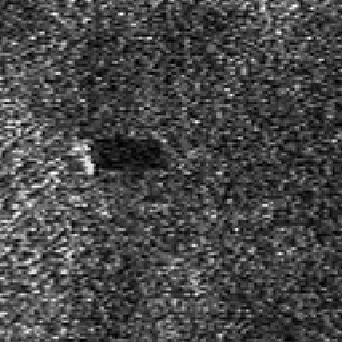}};
\node[anchor=south, inner sep=0] (nonn3) at (\figsepd,\nextfigheightd){\includegraphics[clip,trim=0 15 0 15,width=\figwidthd]{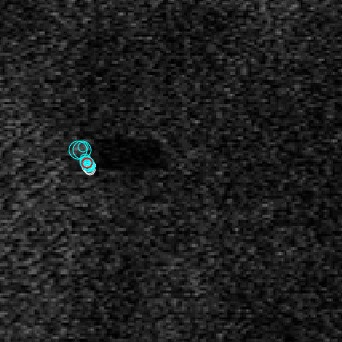}};
\node[anchor=south, inner sep=0] (conv113)  at (2*\figsepd,\nextfigheightd){\includegraphics[clip,trim=0 15 0 15,width=\figwidthd]{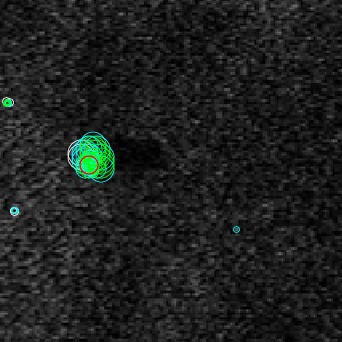}};
\node[anchor=south, inner sep=0] (conv213)  at (3*\figsepd,\nextfigheightd){\includegraphics[clip,trim=0 15 0 15,width=\figwidthd]{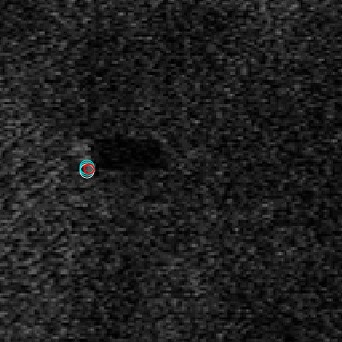}};
\node[anchor=south, inner sep=0] (conv313)  at (4*\figsepd,\nextfigheightd){\includegraphics[clip,trim=0 15 0 15,width=\figwidthd]{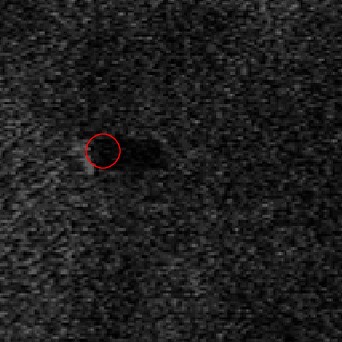}};
\node[anchor=south, inner sep=0] (salicon3)  at (8*\figsepd,\nextfigheightd){\includegraphics[clip,trim=0 15 0 15,width=\figwidthd]{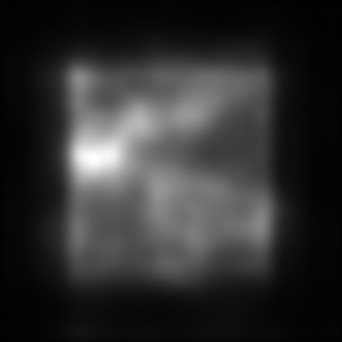}};
\node[anchor=south, inner sep=0] (itti3) at (6*\figsepd,\nextfigheightd){\includegraphics[clip,trim=0 15 0 15,width=\figwidthd]{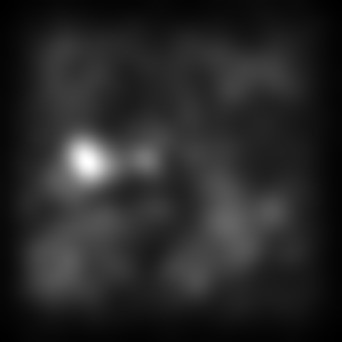}};
\node[anchor=south, inner sep=0] (cohen3) at (5*\figsepd,\nextfigheightd){\includegraphics[clip,trim=0 15 0 15,width=\figwidthd]{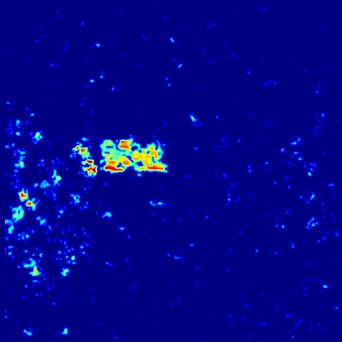}};
\node[anchor=south, inner sep=0] (drfi3) at (7*\figsepd,\nextfigheightd){\includegraphics[clip,trim=0 15 0 15,width=\figwidthd]{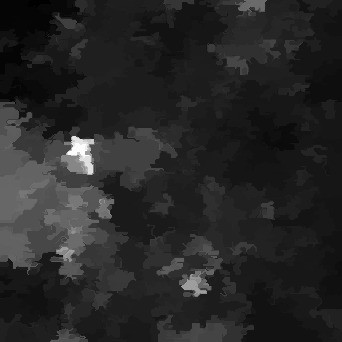}};

\addtolength{\nextfigheightd}{-0.9\figsepd}

\node[anchor=south, inner sep=0] (example6) at (0,\nextfigheightd) {\includegraphics[clip,trim=0 15 0 15,width=\figwidthd]{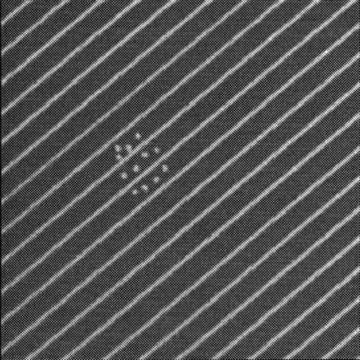}};
\node[anchor=south, inner sep=0] (nonn6) at (\figsepd,\nextfigheightd){\includegraphics[clip,trim=0 15 0 15,width=\figwidthd]{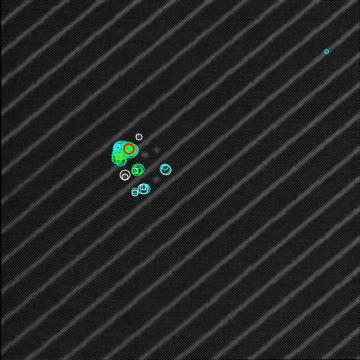}};
\node[anchor=south, inner sep=0] (conv116)  at (2*\figsepd,\nextfigheightd){\includegraphics[clip,trim=0 15 0 15,width=\figwidthd]{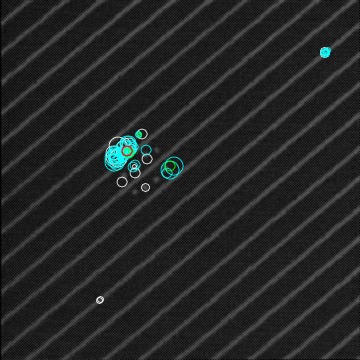}};
\node[anchor=south, inner sep=0] (conv216)  at (3*\figsepd,\nextfigheightd){\includegraphics[clip,trim=0 15 0 15,width=\figwidthd]{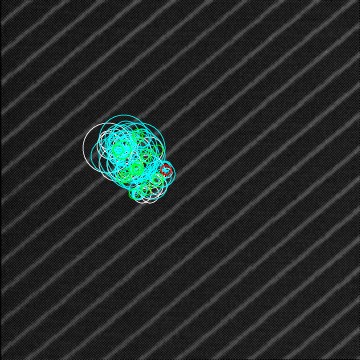}};
\node[anchor=south, inner sep=0] (conv316)  at (4*\figsepd,\nextfigheightd){\includegraphics[clip,trim=0 15 0 15,width=\figwidthd]{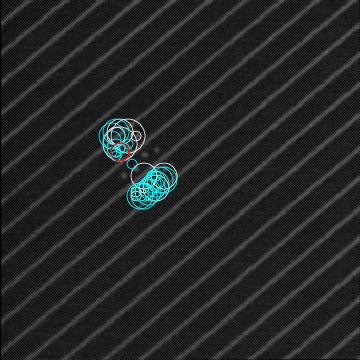}};
\node[anchor=south, inner sep=0] (salicon6)  at (8*\figsepd,\nextfigheightd){\includegraphics[clip,trim=0 15 0 15,width=\figwidthd]{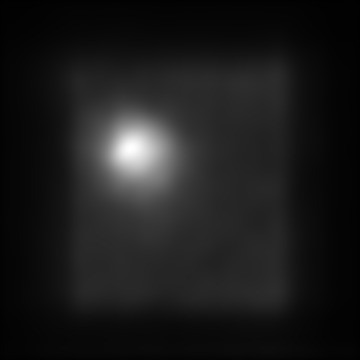}};
\node[anchor=south, inner sep=0] (itti6) at (6*\figsepd,\nextfigheightd){\includegraphics[clip,trim=0 15 0 15,width=\figwidthd]{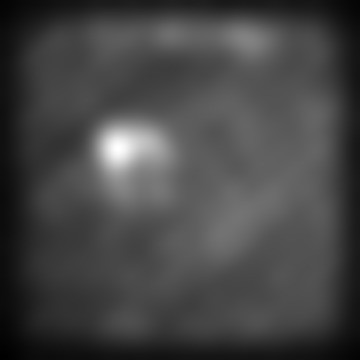}};
\node[anchor=south, inner sep=0] (cohen6) at (5*\figsepd,\nextfigheightd){\includegraphics[clip,trim=0 15 0 15,width=\figwidthd]{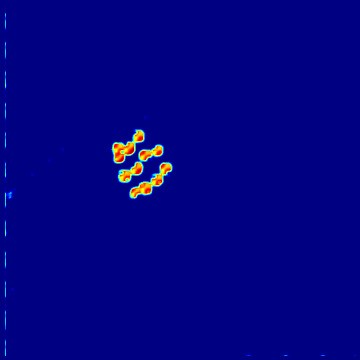}};
\node[anchor=south, inner sep=0] (drfi6) at (7*\figsepd,\nextfigheightd){\includegraphics[clip,trim=0 15 0 15,width=\figwidthd]{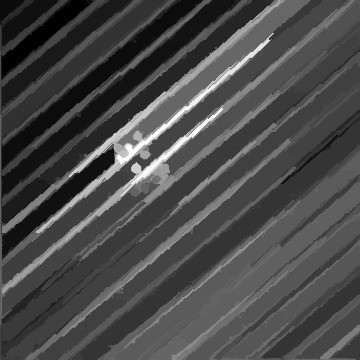}};

  \end{tikzpicture}

  \caption{Detection results on synthetic (top four rows) and real (bottom four rows) images. Detections represented by circles, with radius proportional to detected scale and color to detection strength (NFA). White: weak detection - NFA $\in [10^{-3}, 10^{-2}]$, cyan: mild detection - NFA $\in [10^{-8}, 10^{-3}]$, green: strong detection - NFA $\in [10^{-21}, 10^{-8}]$, and orange: very strong detection - NFA $\le 10^{-21}$. Red: detection with lowest NFA. 
Examples in rows 5th and 6th are from the Toronto dataset~\cite{bruce2006saliency} while  7th and 8th from \cite{mishne2013multiscale} and \cite{tsai1999automated} respectively.}
  
 \label{fig:toy_real}
\end{figure*}

In absence  of a valid test image  database for  anomalies, we used the  most common images proposed in the  literature  (see Fig.~\ref{fig:toy_real}) and we  adopted the following  comparison methodology, that was applied  to  our method and to other four state-of-the-art ones for comparison:

a) \textit{Sanity check:} verifying that for toy examples proposed in the  literature the sole detection is the anomaly;

b) \textit{Theoretical sanity check:}  verify the \textit{a contrario} principle:  "no detection  in  white noise"

c) \textit{Classic  challenging images:}  we  verify the detector power on  classic challenging  images of  the  literature: side scan sonar, textile, mammography and natural images.  In the case of the  mammography where one paper computed  an NFA,  we verify crucially that by computing the  NFA on  the  residual instead of the image, we gain a  huge factor, the  NFA being  divided by eleven orders of magnitude. 
\vspace{.5em}

We tested our proposed anomaly detector on two different input image representations: the basic one, \texttt{pixels}, directly applies the anomaly detection procedure to the residuals obtained from the color channels, and three different variants using as input features extracted at different levels from the VGG network~\cite{simonyan2014very}, namely, very low level~ (\texttt{conv1\_1}), low level~(\texttt{conv2\_1}), and medium level~ (\texttt{conv3\_1}) features. As we shall check the  four detections are similar and  can be fused by a mere pixel union of  all detections.

Existent anomaly detectors are often tuned for specific applications, which probably explains the poor code availability. We compared to Mishne and Cohen~\cite{mishne2013multiscale}, a state-of-the-art anomaly detector with available code, to the salient object detector DRFI~\cite{jiang2013salient} (which is state-of-the-art according to \cite{borji2015salient}), and to the state-of-the-art human gaze predictor SALICON~\cite{huang2015salicon}. 
We also compared to the Itti~\textit{et al.} salient object detector \cite{itti1998model}, which works reasonably well for anomaly detection. All methods produce saliency maps where anomalies have the highest score. Anomalies for Mishne and Cohen are red-colored, while the other methods don't have a threshold for anomalies. 
More results are available in the supplementary materials.

\vspace{.35em}  \noindent \textbf{Synthetic images.}
The proposed method performs well on synthetic examples as shown in Figure~\ref{fig:toy_real}). Some weak false detections are found when using as input features extracted at different layers of the VGG net. All the other compared methods miss some detections. SALICON successfully detects the anomalous density on the fourth example but  misses several anomalies in others or introduces numerous wrong detections. Itti~\etal~method  successfully detects the anomalous color structure in the first example, but fails to detect the other ones.  Mishne and Cohen and DRFI methods do not perform well on any of the five synthetic examples.

\vspace{.35em}  \noindent \textbf{Real images.}
The comparison on real images is more intricate and requires looking in detail to find out whether detections make sense (Figure~\ref{fig:toy_real}). In the garage door  (fourth row), there are two detections that stand out (lens flare and red sign), some others -- less visible -- can be found (door scratches or holes in the brick wall). For our method, the main detections are present in all the variants.
There are also specific anomalies that can be detected only at a given layer of the neural network. For example, %
\texttt{conv1\_1} detects the holes in the brick wall and the gap between the garage door and the wall, in addition to the ones detected with  \texttt{pixels} input. The variants \texttt{conv2\_1} and \texttt{conv3\_1} detect a missing part of a brick in the wall. Saliency methods detect the red sign but not the lens flare.  Mishne and Cohen one only detects the garage door gap.
The second real example is a man walking in front of some trees. Our method detects the man with \texttt{pixels} and \texttt{conv1\_1}. 
DRFI and SALICON detect the man while Mishne and Cohen and Itti~\etal~do not.
The third real example is a radar image showing a mine, while the last example is a defect in a periodic textile. All methods detect the anomalies, with more or less precision. Note that the detection in the top right corner for both \texttt{pixels} and \texttt{conv1\_1} (and only these) correspond to a defect inside the periodic pattern.

\vspace{.5em}  

\begin{figure}[ht]
\includegraphics[clip,trim=0 0 0 60,width=0.32\linewidth]{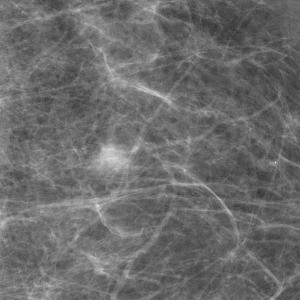}
\includegraphics[clip,trim=0 0 0 60,width=0.32\linewidth]{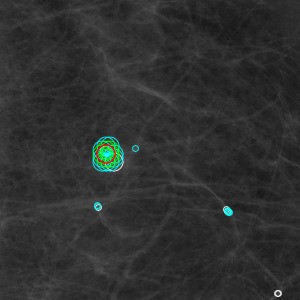}
\includegraphics[clip,trim=0 0 0 60,width=0.32\linewidth]{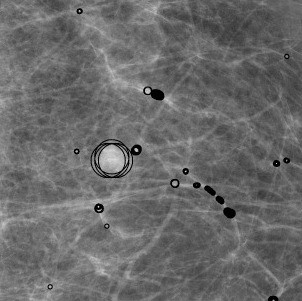}
\caption{The region represented by the large white spot in the left image is a tumor. The proposed self-similarity anomaly detector successfully detects the tumor with a much significant NFA than the one from Grosjean and Moisan~\cite{grosjean2009contrario} (an NFA of $10^{-12}$ versus their reported NFA of $0.15$), while making fewer false detections.}
\label{fig:grosjean}
\end{figure}

\noindent \textbf{Comparison to the \textit{a contrario} method of Grosjean and Moisan~\cite{grosjean2009contrario}.}
This \textit{a contrario} method is designed to detect  spots in colored noise textures, and was applied to the detection of tumors in mammographies.
This detection algorithm is the only other one computing NFAs, and we can directly compare them to ours. The detection results on a real mammography (having a tumor) are shown in Figure~\ref{fig:grosjean}.
With our method the tumor is detected with a much significant NFA (NFA of $10^{-12}$ whereas in~\cite{grosjean2009contrario} NFA of $0.15$). Our  self-similar anomaly detection method shows fewer false detections, actually corresponding to rare events like the crossings of arterials.

\vspace{-.6em}
\section{Conclusion}
\label{sec:conclusions}
\vspace{-.5em}

We have shown that anomalies are easier detected on the residual image, computed by removing the self-similar component, and then performing hypothesis testing. 
It is reassuring to see that our method finds all anomalies  proposed  in  the  literature with very low  NFA.  
In addition, we have experimentally shown that the method verifies the non-accidentalness principle: no anomalies are detected in white noise.  
We plan to build  a database of test  images with anomalies  to run extensive validation and comparison. 
We also plan to extend the method to videos, by analyzing anomalies in the  motion field.

\vspace{-.5em}

\bibliographystyle{IEEEbib}
\bibliography{anomaly_detection_icip}

\end{document}